%% file: main.tex
\documentclass[a4paper,fleqn]{cas-dc}
\usepackage[numbers,sort&compress]{natbib}
\usepackage{booktabs}
\usepackage{tikz}
\usetikzlibrary{positioning,arrows.meta,fit,calc}
\usepackage{placeins}
\newcommand{\Description}[1]{}
\ExplSyntaxOn
\cs_set:Npn \__first_footerline: {\small\rmfamily\itshape Preprint}
\cs_set:Npn \__first_foot: {\parbox[t]{\textwidth}{\rule{\textwidth}{.2pt}\\\__first_footerline: \hfill Page~\thepage}}
\cs_set:Npn \__cas_foot: {\parbox[t]{\textwidth}{\rule{\textwidth}{.2pt}\\\sffamily\small\__first_footerline: \hfill Page~\thepage}}
\ExplSyntaxOff
\shorttitle{Author-Conditioned Style Transfer and AI-Text Detection}
\begin{document}

\title[mode=title,size=\LARGE,weight=\bfseries]{Who Can You Hide Behind? Author-Conditioned Style Transfer and the Evasion of a Commercial AI-Text Detector}

\author[1]{Mohd Ruhul Ameen}
\ead{ruhul.cse@vu.edu.bd}
\author[2]{M. Saifuzzaman Rafat}
\ead{saif.rafat@gmail.com}
\author[2]{Akif Islam}[orcid=0009-0001-5300-5310]
\ead{iamakifislam@gmail.com}
\affiliation[1]{organization={Varendra University}, city={Rajshahi}, country={Bangladesh}}
\affiliation[2]{organization={University of Rajshahi}, city={Rajshahi}, country={Bangladesh}}

\begin{abstract}
AI-text detectors can be sensitive to changes in writing style. We examined this sensitivity by rewriting 991 AI-generated answers with Gemini, using prompts that included a writing sample from one of 55 named authors. We scored each original and its rewrite with the QuillBot AI detector. The median score fell from 100\% to 0\%, and the mean fell from 67.9\% to 11.2\%. Of the rewrites, 66\% received a score of exactly 0\%. Among the 699 originals scored at least 50\%, 71\% of rewrites scored below 10\% (paired Wilcoxon signed-rank test, $p < 10^{-100}$).

Scores varied across the authors used in the prompts. All rewrites associated with 10 authors scored 0\%, whereas the mean score remained above 50\% for six authors (Kruskal--Wallis test, $p < 10^{-80}$). Greater reuse of five-word sequences from the author sample was associated with lower scores, although rewrites with less than 5\% reuse still averaged 15.2\%. These results describe one detector and one rewriting system, not detectors in general. The study lacks a generic-rewrite baseline, did not measure preservation of meaning, and produced rewrites that were substantially longer than the originals.
\end{abstract}


\maketitle

\section{Introduction}

\begin{figure}[pos=!htbp]
  \centering
  \includegraphics[width=\columnwidth]{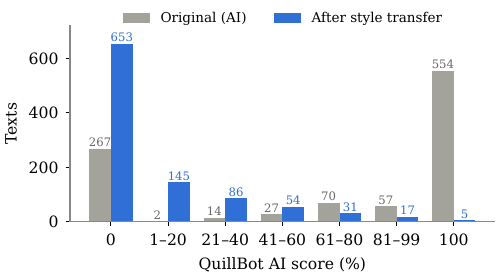}
  \caption{QuillBot AI scores for 991 AI-generated texts before (grey) and after (blue) author-conditioned style transfer. The mass moves from 100\% to 0\%. Numbers above bars are text counts.}
  \Description{Grouped bar chart over seven score bins. Originals: 267 at 0 percent and 554 at 100 percent. Rewrites: 653 at 0 percent and only 5 at 100 percent.}
  \label{fig:teaser}
\end{figure}

AI-text detectors are used in settings where their judgments can have consequences for writers. It is therefore important to understand when their scores change and what those changes mean. Earlier studies have shown that paraphrasing can reduce detection scores. We examine a related question: what happens when an AI-generated answer is rewritten using a sample of a named author's writing, and do the results vary across authors? Figure~\ref{fig:teaser} summarizes the observed change.

We used a chat interface to rewrite 991 answers, each in response to a prompt containing an excerpt from one of 55 writers. The writers included essayists, journalists, scientists and novelists.

We contribute a paired analysis of detector scores before and after rewriting, a comparison across the named authors used in the prompts, and an examination of text reuse and length. This is an observational analysis of one detector and one rewriting system. As discussed in Section~\ref{sec:discussion}, it does not isolate the effect of author identity from other features of the prompts or establish that the rewrites preserve the original meaning.

\section{Related Work}

\paragraph{Detection methods and evaluation.} Machine-generated text detection includes statistical methods, trained classifiers and watermarking; a recent survey reviews these approaches and their evaluation challenges \citep{wu-etal-2025-survey}. DetectGPT examines probability curvature, while Binoculars compares scores from two language models \citep{mitchell2023detectgpt,hans2024binoculars}. Watermarking embeds a detectable signal during generation \citep{kirchenbauer2023watermark}. These approaches differ from the commercial detector examined here. The M4GT benchmark spans languages, domains and generators and illustrates the importance of testing beyond a single setting \citep{wang-etal-2024-m4gt}.

\paragraph{Rewriting and detection.} Krishna et al. showed that paraphrasing can reduce detection rates and examined retrieval as a defense \citep{krishna2023paraphrasing}. Sadasivan et al. studied recursive paraphrasing against multiple detector families \citep{sadasivan2023can}, and Cheng et al. used detector feedback to guide paraphrasing \citep{cheng2025adversarial}. Other studies have tested attacks involving editing, prompting and paraphrasing, including small adversarial changes and combinations of detectors \citep{wang-etal-2024-stumbling,zhou-etal-2024-humanizing,shportko-verbitsky-2025-paraphrasing}. A study of commercial humanizers and paraphrasing tools also examined their effects on detection and the meaning of the resulting text \citep{masrour-etal-2025-damage}. In a separate practical evaluation, detector performance varied across unseen data and adversarial prompts \citep{tufts-etal-2025-practical}.

\paragraph{Style and authorship.} Changes in writing style and text complexity can affect machine-text detection \citep{doughman-etal-2025-exploring}; other work has proposed style-aware detection methods \citep{wu-etal-2025-moses}. Authorship obfuscation and mimicry are related but distinct tasks: StyleRemix changes selected stylistic features to obscure authorship \citep{fisher-etal-2024-styleremix}, while Nguyen et al. examine language models in authorship obfuscation, mimicry and verification \citep{nguyen-etal-2025-unraveling}.

\paragraph{Prompt-based evasion.} Lu et al. developed SICO, which optimizes human-written in-context examples to guide a model toward text that evades detectors \citep{lu2024sico}. Our study instead examines a manual chat workflow using excerpts attributed to named writers and compares the resulting scores across those writers. These differences do not establish that author identity itself caused the observed changes.

\section{Method}

\begin{figure}[pos=!htbp]
  \centering
  \begin{tikzpicture}[
      font=\footnotesize,
      box/.style={draw=black!45, rounded corners=2pt, fill=#1, align=center, minimum height=10mm, text width=21mm, inner sep=2pt},
      box/.default=black!4,
      arr/.style={-{Stealth[length=1.6mm]}, draw=black!60, line width=0.5pt},
      node distance=5mm and 4mm]
    \node[box] (src) {AI answer\\{\scriptsize 991 texts}};
    \node[box=blue!6, right=of src] (prompt) {Author excerpt\\+ style\\instruction\\{\scriptsize 55 authors}};
    \node[box=blue!6, right=of prompt] (gem) {Gemini\\{\scriptsize chat, manual}};
    \node[box=blue!12, below=of gem] (rw) {Rewrite};
    \node[box, below=of src] (det1) {QuillBot\\score \emph{before}};
    \node[box, below=of prompt] (det2) {QuillBot\\score \emph{after}};
    \coordinate (top) at ($(prompt.north)+(0,4mm)$);
    \draw[arr] (prompt) -- (gem);
    \draw[arr] (gem) -- (rw);
    \draw[arr] (rw) -- (det2);
    \draw[arr] (src) -- (det1);
    \draw[arr, dashed] (src.north) -- (src.north |- top) -- node[above, font=\scriptsize] {content to rewrite} (gem.north |- top) -- (gem.north);
  \end{tikzpicture}
  \caption{Data collection. Each AI answer was rewritten once in one author's style; the original and the rewrite were each scored once.}
  \Description{Flow diagram: AI answer goes to a prompt with an author excerpt, then to Gemini, which produces a rewrite. The original and the rewrite are each scored by QuillBot.}
  \label{fig:pipeline}
\end{figure}
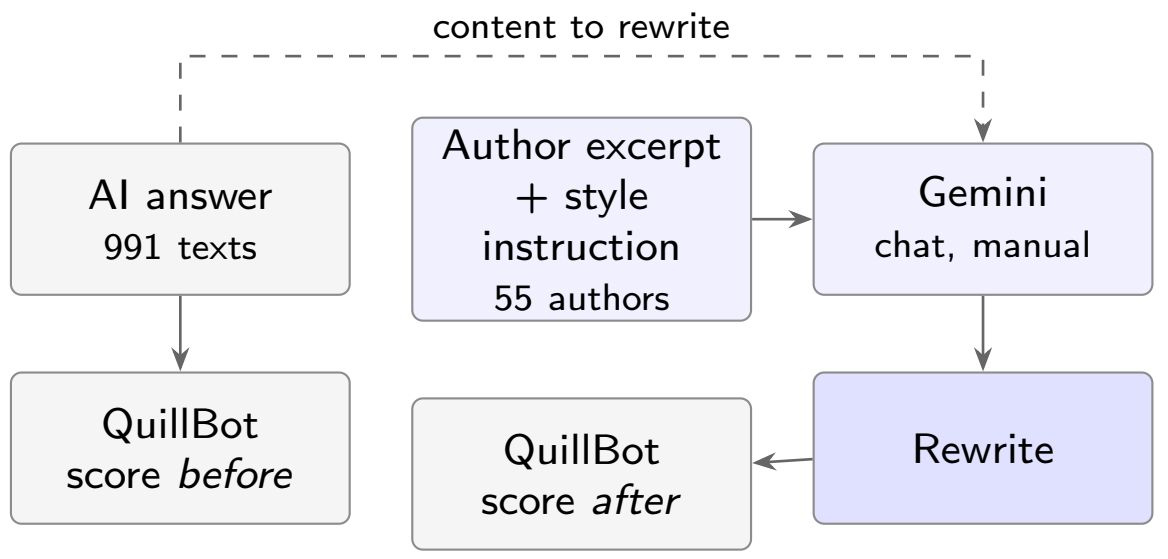

Each observation consists of an AI-generated answer, one rewrite conditioned on an author sample, and a detector score for each text (Figure~\ref{fig:pipeline}).

\paragraph{Source texts.} The original texts are AI-generated answers to everyday questions, including questions about wine pairing, budgeting, job searches and cooking. Their median length was 92 words. The source dataset, its licence and the sampling procedure were not documented in the available study materials, limiting reproducibility.

\paragraph{Authors.} We used 55 authors, 5 to 31 texts each (median 18). The set spans essayists, journalists, scientists, AI researchers and novelists. The full list is in Table~\ref{tab:authors}.

\paragraph{Prompt.} The prompt names the author, states that the sample is their writing, includes one excerpt of their published work (median prompt length 436 words), and asks the model to learn the style and rewrite the provided content in it. The source text was supplied separately. The prompt was refined during collection: 25 prompts add an instruction to keep the word count close to the original, and 23 tell the model to avoid a repeated opening (``I was talking to a friend\ldots''). For 18 authors more than one prompt variant was used.

\paragraph{Generation.} Rewrites were produced manually through the Gemini chat interface. The available materials did not document the model version, collection dates or whether a fresh chat was used for each rewrite.

\paragraph{Detection.} Each original and rewrite was scored once with the QuillBot AI detector. We report its percentage score as returned by the tool; it should not be interpreted as a calibrated probability of AI authorship. The available materials did not document the collection dates or detector settings.

\paragraph{Cleaning.} The collection sheet holds 1,099 data rows. We dropped 99 empty rows, 6 rows with a missing or non-numeric score, 2 rows with no author, and 1 row with an impossible score (1000\%). The resulting analysis contained 991 pairs. One author name was misspelled in the sheet (``George Monbio''); we report it as George Monbiot. Table~\ref{tab:cleaning} lists every change.

\paragraph{Measures.} We report the before and after scores, the share of rewrites at 0\%, and two text measures. \emph{Length ratio} is rewrite words divided by original words. \emph{Sample reuse} is the fraction of the rewrite's word 5-grams that also appear in the prompt's author sample. We tested the paired change with a Wilcoxon signed-rank test and differences across authors (those with at least 5 texts) with a Kruskal--Wallis test. These tests assess associations within the collected sample; they do not establish causation.

\section{Results}

\subsection{Overall Effect}

Author-conditioned rewriting cut the mean detector score from 67.9\% to 11.2\% and the median from 100\% to 0\% ($n = 991$; Wilcoxon signed-rank, $p < 10^{-100}$). Of all rewrites, 653 (66\%) scored exactly 0\% and 82 (8\%) scored 50\% or higher.

Many originals received low scores before rewriting: 269 scored below 20\%. We therefore also report results for the subset with original scores of at least 50\% (Table~\ref{tab:overall}).

\begin{table}[pos=!htbp]
\caption{Detector scores before and after rewriting.}
\label{tab:overall}
\footnotesize
\setlength{\tabcolsep}{2.2pt}
\begin{tabular}{lrrrrrr}
\toprule
Subset & $n$ & Before & After & $=0\%$ & $<10\%$ & $\geq 50\%$ \\
\midrule
All pairs & 991 & 67.9\% & 11.2\% & 653 (66\%) & 708 (71\%) & 82 (8\%) \\
Orig.\ $\geq 50\%$ & 699 & 95.1\% & 12.0\% & 454 (65\%) & 494 (71\%) & 67 (10\%) \\
\bottomrule
\end{tabular}
\end{table}

\begin{figure}[pos=!htbp]
  \centering
  \includegraphics[width=\columnwidth]{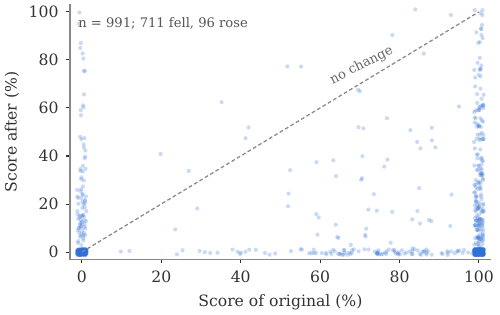}
  \caption{Paired scores per text (small jitter added). Points below the dashed line fell after rewriting. Most rewrites sit on the floor at 0\% regardless of the original score.}
  \Description{Scatter plot of score after against score before. Most points lie along the bottom edge at 0 percent; a column of points sits at an original score of 100 percent.}
  \label{fig:scatter}
\end{figure}

Figure~\ref{fig:scatter} shows the paired scores: 711 fell, 184 were unchanged and 96 rose. In 86 of the 96 pairs with an increase, the original had scored below 20\%.

\subsection{Variation Across Authors}

\begin{figure*}[pos=!tp]
  \centering
  \includegraphics[width=\textwidth]{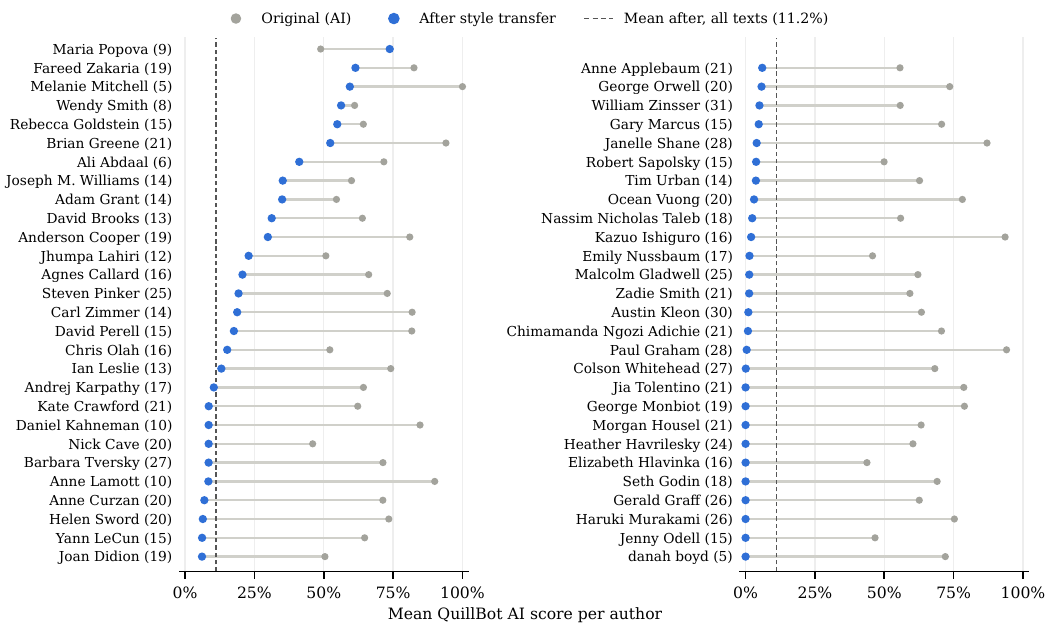}
  \caption{Mean QuillBot AI score before (grey) and after (blue) rewriting, per imitated author, sorted by the after score (read the left panel, then the right). Number of texts in brackets. Ten authors reach 0\% on every rewrite; six stay above 50\%.}
  \Description{Two-panel dumbbell chart of 55 authors. Top-left authors such as Maria Popova and Fareed Zakaria keep high after-scores; the right panel shows authors whose after-scores are at or near 0 percent.}
  \label{fig:authors}
\end{figure*}

Scores differed across the authors represented in the prompts (Kruskal--Wallis on rewrite scores across 55 authors, $H = 541$, $p < 10^{-80}$; Figure~\ref{fig:authors}). For 10 authors every rewrite scored 0\%, including Jia Tolentino, George Monbiot, Morgan Housel, Seth Godin and Haruki Murakami. For 6 authors the mean rewrite score stayed above 50\%: Maria Popova (73.8\%), Fareed Zakaria (61.4\%), Melanie Mitchell (59.4\%), Wendy Smith (56.2\%), Rebecca Goldstein (54.9\%) and Brian Greene (52.3\%). None of Popova's, Zakaria's, Greene's or Mitchell's rewrites reached 0\%.

\input{per_author_table}

\subsection{Reuse of the Author Sample}

Some rewrites reused five-word sequences from the human-written sample included in the prompt. Greater measured reuse was associated with a lower score (Spearman $\rho = -0.30$, $p < 10^{-20}$). The association does not establish that copying caused the lower scores. Rewrites with less than 5\% measured reuse averaged 15.2\%, compared with an overall original-text mean of 67.9\% (Table~\ref{tab:reuse}).

\begin{table}[pos=!htbp]
\caption{Rewrite score by reuse of the author sample (share of the rewrite's word 5-grams found in the sample).}
\label{tab:reuse}
\footnotesize
\begin{tabular}{lrrr}
\toprule
5-grams reused & $n$ & Mean after & After $=0\%$ \\
\midrule
Under 5\% & 663 & 15.2\% & 376 (57\%) \\
5\% to 20\% & 213 & 3.9\% & 167 (78\%) \\
Over 20\% & 115 & 1.5\% & 110 (96\%) \\
\bottomrule
\end{tabular}
\end{table}

\begin{figure}[pos=!htbp]
  \centering
  \includegraphics[width=\columnwidth]{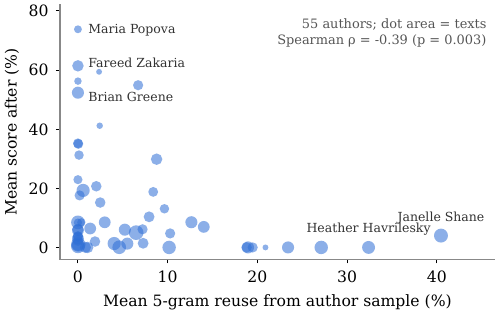}
  \caption{Author-level reuse of the human sample against mean score after rewriting. Authors with high reuse all sit near 0\%; the authors that stay detectable reuse almost nothing.}
  \Description{Scatter of 55 authors. Authors with more than 15 percent reuse all have after-scores near 0. Maria Popova, Fareed Zakaria and Brian Greene sit at near-zero reuse with after-scores above 50 percent.}
  \label{fig:reuse}
\end{figure}

Reuse also clusters by author (Figure~\ref{fig:reuse}; author-level Spearman $\rho = -0.39$, $p = 0.003$). Median reuse was 42\% for Janelle Shane and 38\% for Heather Havrilesky, but at most 1\% for all six authors whose rewrites stayed above 50\%.

\subsection{Length}

Rewrites were about four times longer than originals (median 92 $\rightarrow$ 428 words; ratio quartiles 2.9, 4.1, 5.8). Rewrite length itself was not associated with the score ($\rho = -0.01$, $p = 0.75$), and the length ratio only weakly ($\rho = -0.11$).

\section{Discussion}
\label{sec:discussion}

The detector scores differed substantially across the author-conditioned prompts. Because author samples, source texts and prompt variants were not controlled independently, the analysis cannot attribute these differences to author identity or writing style alone.

\paragraph{Possible explanations.} The authors whose rewrites stayed detectable (Popova, Zakaria, Greene, Mitchell, Goldstein) write mostly explanatory prose about ideas, science or policy. The authors that reach 0\% more often write personal, anecdotal or conversational prose (Tolentino, Havrilesky, Godin, Housel). One hypothesis is that the detector treats measured expository prose as AI-like regardless of origin. We have not tested this. A test would score each author's own human-written excerpt with the same detector and measure stylistic features such as sentence-length variance, first-person use and anecdote frequency.

\paragraph{Threats to validity.} A score of 0\% means only that QuillBot did not flag the text; it does not establish that the text reads as human or preserves the original meaning. The rewrites were substantially longer and usually reframed the content as an essay, and meaning preservation was not measured. Reuse of the human sample is another potential confound: rewrites with more reuse tended to score lower, while our five-gram measure captures near-verbatim overlap rather than close paraphrase.

The study also lacks a generic ``make this sound human'' baseline and a baseline using an anonymous human sample. We therefore cannot separate the influence of a named author from that of rewriting or providing a sample. The prompt changed during collection, and 18 authors had more than one variant, so author and prompt version are partly confounded. Finally, the results cover one detector, one rewriting system and one manually collected score per text. Both commercial detectors and Gemini can change over time, which limits reproducibility and generalization.

\section{Limitations and Ethics}

This study identifies a weakness in one detector under the conditions examined. The findings should not be generalized to all detectors. Reporting such limitations can help readers interpret detector scores with appropriate caution.

The prompts include excerpts attributed to published authors. Any data release should respect the rights associated with those excerpts; access and redistribution arrangements have not been documented in the available study materials.

The rewrites put words in the voice of real, named writers. They are not presented as the authors' work, and readers should not attribute them to those authors.

\section{Conclusion}

Across 991 pairs, author-conditioned rewriting was associated with a decrease in the detector's median score from 100\% to 0\%. Scores varied across authors represented in the prompts, and greater sample reuse was associated with lower scores. The design does not establish why these patterns occurred.

The next study should add a generic-rewrite baseline, several detectors and generators, a meaning-preservation measure, and scores for each author's own human writing. That design would test whether detectors confuse expository style with machine origin.

\appendix

\section{Data Cleaning Log}

\begin{table}[pos=!htbp]
\caption{Rows changed or removed during cleaning (sheet row numbers).}
\label{tab:cleaning}
\footnotesize
\setlength{\tabcolsep}{3pt}
\begin{tabular}{p{0.25\columnwidth}p{0.34\columnwidth}p{0.22\columnwidth}}
\toprule
Sheet row & Problem & Action \\
\midrule
99 rows & Empty & Dropped \\
100, 192, 290, 918, 1000 & Rewrite score missing & Dropped \\
110 & Score cell holds pasted text, no author & Dropped \\
535, 902 & No author name & Dropped \\
904 & Original score recorded as 1000\% & Dropped \\
222--240 & Author spelled ``George Monbio'' & Reported as George Monbiot \\
\bottomrule
\end{tabular}
\end{table}

\bibliographystyle{IEEEtran}
\bibliography{references}

\end{document}

%% file: per_author_table.tex
\begin{table*}[pos=!tp]
\caption{Per-author results. $n$: texts. At 0\%: share of rewrites scored 0\%. Reuse: median share of the rewrite's word 5-grams found in the author sample. Len.: median length ratio (rewrite/original). Mean scores before and after are in Figure~\ref{fig:authors}.}
\label{tab:authors}
\footnotesize
\begin{tabular}{lrrrr}
\toprule
Author & $n$ & At 0\% & Reuse & Len.\\
\midrule
Adam Grant & 14 & 7\% & 0\% & 6.4$\times$ \\
Agnes Callard & 16 & 31\% & 2\% & 4.4$\times$ \\
Ali Abdaal & 6 & 17\% & 1\% & 6.2$\times$ \\
Anderson Cooper & 19 & 16\% & 0\% & 2.7$\times$ \\
Andrej Karpathy & 17 & 24\% & 8\% & 5.9$\times$ \\
Anne Applebaum & 21 & 67\% & 5\% & 3.5$\times$ \\
Anne Curzan & 20 & 75\% & 16\% & 4.0$\times$ \\
Anne Lamott & 10 & 80\% & 0\% & 4.0$\times$ \\
Austin Kleon & 30 & 93\% & 0\% & 6.6$\times$ \\
Barbara Tversky & 27 & 59\% & 0\% & 2.7$\times$ \\
Brian Greene & 21 & 0\% & 0\% & 2.6$\times$ \\
Carl Zimmer & 14 & 43\% & 7\% & 3.2$\times$ \\
Chimamanda Ngozi Adichie & 21 & 90\% & 0\% & 4.4$\times$ \\
Chris Olah & 16 & 50\% & 3\% & 2.7$\times$ \\
Colson Whitehead & 27 & 96\% & 0\% & 4.1$\times$ \\
Daniel Kahneman & 10 & 70\% & 0\% & 2.3$\times$ \\
David Brooks & 13 & 8\% & 0\% & 6.0$\times$ \\
David Perell & 15 & 27\% & 0\% & 6.5$\times$ \\
Elizabeth Hlavinka & 16 & 100\% & 18\% & 5.1$\times$ \\
Emily Nussbaum & 17 & 94\% & 7\% & 6.9$\times$ \\
Fareed Zakaria & 19 & 0\% & 0\% & 2.6$\times$ \\
Gary Marcus & 15 & 73\% & 12\% & 4.7$\times$ \\
George Monbiot & 19 & 100\% & 0\% & 3.8$\times$ \\
George Orwell & 20 & 75\% & 0\% & 4.3$\times$ \\
Gerald Graff & 26 & 100\% & 28\% & 4.6$\times$ \\
Haruki Murakami & 26 & 100\% & 12\% & 6.0$\times$ \\
Heather Havrilesky & 24 & 100\% & 38\% & 7.8$\times$ \\
Helen Sword & 20 & 60\% & 0\% & 4.4$\times$ \\
\bottomrule
\end{tabular}\hfill
\begin{tabular}{lrrrr}
\toprule
Author & $n$ & At 0\% & Reuse & Len.\\
\midrule
Ian Leslie & 13 & 38\% & 11\% & 4.2$\times$ \\
Janelle Shane & 28 & 82\% & 42\% & 3.0$\times$ \\
Jenny Odell & 15 & 100\% & 20\% & 6.4$\times$ \\
Jhumpa Lahiri & 12 & 33\% & 0\% & 4.3$\times$ \\
Jia Tolentino & 21 & 100\% & 19\% & 2.8$\times$ \\
Joan Didion & 19 & 68\% & 0\% & 2.3$\times$ \\
Joseph M. Williams & 14 & 36\% & 0\% & 1.8$\times$ \\
Kate Crawford & 21 & 71\% & 14\% & 3.6$\times$ \\
Kazuo Ishiguro & 16 & 81\% & 2\% & 3.5$\times$ \\
Malcolm Gladwell & 25 & 84\% & 4\% & 6.5$\times$ \\
Maria Popova & 9 & 0\% & 0\% & 4.9$\times$ \\
Melanie Mitchell & 5 & 0\% & 1\% & 4.8$\times$ \\
Morgan Housel & 21 & 100\% & 24\% & 3.4$\times$ \\
Nassim Nicholas Taleb & 18 & 78\% & 0\% & 2.5$\times$ \\
Nick Cave & 20 & 45\% & 2\% & 3.7$\times$ \\
Ocean Vuong & 20 & 65\% & 0\% & 3.9$\times$ \\
Paul Graham & 28 & 96\% & 0\% & 5.2$\times$ \\
Rebecca Goldstein & 15 & 7\% & 0\% & 3.8$\times$ \\
Robert Sapolsky & 15 & 67\% & 0\% & 4.7$\times$ \\
Seth Godin & 18 & 100\% & 1\% & 4.2$\times$ \\
Steven Pinker & 25 & 12\% & 1\% & 3.7$\times$ \\
Tim Urban & 14 & 79\% & 0\% & 2.8$\times$ \\
Wendy Smith & 8 & 0\% & 0\% & 6.3$\times$ \\
William Zinsser & 31 & 81\% & 0\% & 4.8$\times$ \\
Yann LeCun & 15 & 80\% & 8\% & 3.8$\times$ \\
Zadie Smith & 21 & 86\% & 6\% & 5.4$\times$ \\
danah boyd & 5 & 100\% & 23\% & 3.8$\times$ \\
\bottomrule
\end{tabular}
\end{table*}